\documentclass[sigconf]{acmart}

\AtBeginDocument{%
  \providecommand\BibTeX{{%
    \normalfont B\kern-0.5em{\scshape i\kern-0.25em b}\kern-0.8em\TeX}}}

\setcopyright{acmcopyright}
\copyrightyear{2020}
\acmYear{2020}

\acmConference[CIKM '20]{CIKM '20: ACM Conference on Information and Knowledge Management}{October 19--23, 2020}{Galway, Ireland}
\acmBooktitle{CIKM '20: ACM Conference on Information and Knowledge Management,
  October 19--23, 2020, Galway, Ireland}

\usepackage{multirow}
\usepackage[ruled,vlined]{algorithm2e}
\newcommand\footnoteref[1]{\protected@xdef\@thefnmark{\ref{#1}}\@footnotemark}
\makeatother
\usepackage{siunitx}
\usepackage{tikz}
\usepackage{pgfplots}
\pgfplotsset{width=7cm,compat=1.3}
\usepackage[bottom]{footmisc}
\usepackage{textpos}
\usepackage{hyperref}
\usepackage{subfig}
\usepackage{graphicx}

\begin{document}

\title{METEOR: Learning Memory and Time Efficient Representations from Multi-modal Data Streams}
\titlenote{Accepted as a conference paper at CIKM 2020}
\author{Amila Silva, Shanika Karunasekera, Christopher Leckie and Ling Luo}
\affiliation{School of Computing and Information Systems\\ The University of Melbourne \\Parkville, Victoria, Australia}
\email{{amila.silva@student., karus@, caleckie@, ling.luo@}unimelb.edu.au}
\renewcommand{\shortauthors}{Silva A., Karunasekera S., Leckie C., and Luo L.}

\begin{abstract}
Many learning tasks involve multi-modal data streams, where continuous data from different modes convey a comprehensive description about objects. A major challenge in this context is how to efficiently interpret multi-modal information in complex environments. This has motivated numerous studies on learning unsupervised representations from multi-modal data streams. These studies aim to understand higher-level contextual information (e.g., a Twitter message) by jointly learning embeddings for the lower-level semantic units in different modalities (e.g., text, user, and location of a Twitter message). However, these methods directly associate each low-level semantic unit with a continuous embedding vector, which results in high memory requirements. Hence, deploying and continuously learning such models in low-memory devices (e.g., mobile devices) becomes a problem. To address this problem, we present \textsc{METEOR}, a novel \textbf{ME}mory and \textbf{T}ime \textbf{E}fficient \textbf{O}nline \textbf{R}epresentation learning technique, which: (1) learns compact representations for multi-modal data by sharing parameters within semantically meaningful groups and preserves the domain-agnostic semantics; (2) can be accelerated using parallel processes to accommodate different stream rates while capturing the temporal changes of the units; and (3) can be easily extended to capture implicit/explicit external knowledge related to multi-modal data streams. We evaluate \textsc{METEOR} using two types of multi-modal data streams (i.e., social media streams and shopping transaction streams) to demonstrate its ability to adapt to different domains. Our results show that \textsc{METEOR} preserves the quality of the representations while reducing memory usage by around 80\% compared to the conventional memory-intensive embeddings.
\end{abstract}

\maketitle



\section{Introduction}
Unsupervised representation learning~\cite{zhang2018network,bengio2013representation} has recently become a rapidly growing direction in machine learning due to its ability to: (1) exploit the availability of unlabeled data; and (2) associate the underlying factors behind data effortlessly. For a given domain, conventional representational learning techniques learn low-dimensional vectors (i.e., embeddings) for low-level data units (e.g., words in a language) such that these representations capture the underlying semantics of the particular domain in a task-independent manner. For example, the language modelling techniques in~\cite{mikolov2013distributed,kim2016character} learn embeddings for words or characters by considering them as the low-level units\footnote{Such low-level units in different domains are referred to as ``units'' in this paper.} of a language, from which higher-level structures (e.g., phrases and sentences) can be constructed and understood. Subsequently, these representations serve as features to solve different application-level problems. Lately, it has been empirically proven that these representations yield a significant performance boost for many downstream tasks in domains such as Natural Language Processing~\cite{bojanowski2017enriching,mikolov2013distributed} and Computer Vision~\cite{he2016deep,simonyan2014very}.

\begin{figure}[b]
  \centering
  \includegraphics[width=.9\linewidth]{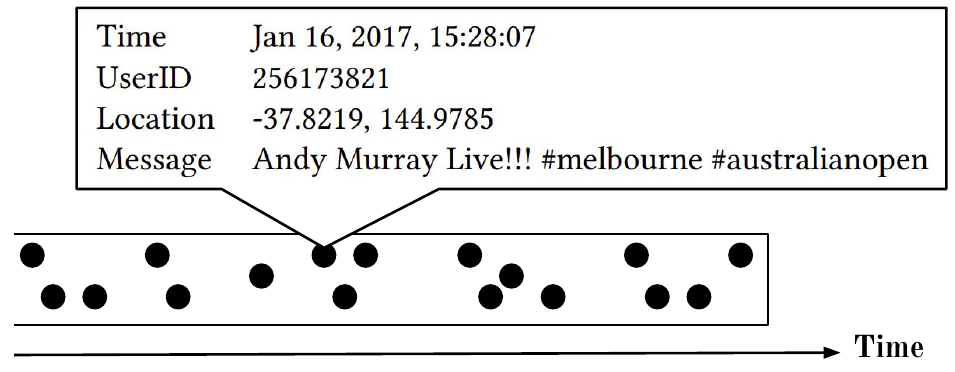} 
  \vspace{-2mm}
\caption{Geo-tagged Twitter Stream as an example of a  multi-modal data stream}
\label{fig:1}
\vspace{-4mm}
\end{figure}

This paper focuses on representation learning techniques for multi-modal data streams: techniques to learn representations from records with different types of low-level units (i.e., attributes) in an online fashion to capture temporal changes of the units while preserving the relationships between different modalities. For example, a geo-tagged Twitter stream (see Figure~\ref{fig:1}) is such a multi-modal data stream, in which each record has multiple types of attributes, such as its location, user, timestamp, and text content. Although there has been previous work~\cite{silva2019ustar,zhang2017react} on this problem, our work aims to address the following research gaps in the literature.

\textbf{Research Gaps. }First, conventional embedding learning techniques are memory-intensive as they assign an independent embedding vector to each low-level unit. If such techniques are extended to embed attributes with a growing number of distinct low-level units, such as users and words, the amount of memory required to store all the embedding vectors becomes a major overhead. For example, consider a system that learns embeddings using a shopping transaction record stream, which consists of two types of low-level units: (1) 1 million distinct users; and (2) 1 million different products. If a 300-dimensional vector is assigned to each unit, the total memory requirement for the embeddings becomes\footnote{Here we assume that the values in the embeddings are represented as single-precision float numbers (4 bytes per value).} $(1,000,000+1,000,000) (units)\times 300 (dimensions)\times 4 (bytes) \approx 2.4 GB$. Moreover, some multi-modal streams can have more than two modalities. If we consider a geo-tagged Twitter stream as an example, there are four modalities, namely: location, text, user, and timestamp. Hence, the total memory required for such a stream can be substantially higher than the example above. Thus, the application of conventional embedding learning approaches for multi-modal streams becomes problematic, particularly on limited memory platforms. There are several previous works (compared in Section~\ref{sec:related_works}) on learning compact memory-efficient representations. However, almost all these methods are not well suited to data streams~\cite{chen2018learning,shu2017compressing}, or they are specific to a particular domain (e.g., the technique proposed in~\cite{sennrich2015neural} is specific to natural language modelling). Thus, this paper proposes \textbf{\textit{a domain-agnostic and memory-efficient representation learning technique to work with data streams}}.

Second, the processing time per record of online learning techniques should meet at least the rate of the stream (i.e., 1/average records per unit time) to update models in a timely manner. Parallel-processing architectures can be used to meet this requirement when working with data streams. However, this problem is not well studied in the context of online representation learning, possibly due to the memory-intensive nature of conventional embeddings. With memory-efficient representations, it is feasible to adopt parallel-processing to reduce the time complexity of online representation learning techniques. 
We propose \textbf{\textit{a decomposable objective function to learn memory-efficient representations from streams, which allows the flexibility to assign more parallel processes (with different memory capacities) for computationally expensive steps in \textsc{METEOR}}}.

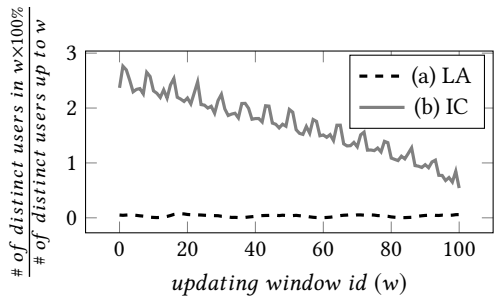
\begin{figure}[t]
\begin{tikzpicture}
\begin{axis}[
    width=7cm,
    height=4cm,
    ylabel={$\frac{\#\; of\;distinct\; users\; in\;w\times100\%}{\#\; of\;distinct\; users\; up\;to\;w}$},
    xlabel={$updating\;window\;id\; (w)$},
]
\addplot [dashed, black,very thick] table [x={id}, y={ny}] {graph_1.dat};
\addlegendentry{(a) LA}
\addplot [gray,very thick] table [x={id}, y={ic}] {graph_1.dat};
\addlegendentry{(b) IC}
\end{axis}
\end{tikzpicture}
\caption{The fraction of different users that appear within an updating window of length $\Delta W$ compared to the total users seen up to the particular window: (a) LA, a Geo-tagged Twitter Stream ($\Delta W=1\;hour$); and (b) IC, a Shopping Transaction Stream ($\Delta W=1\;day$). More details about LA and IC are provided in Section~\ref{sec:experiments}}\label{fig:online_frequency}
\vspace{-4mm}
\end{figure}

Third, some of the attributes in multi-modal data streams show specific relationships or behaviours (either explicit or implicit). For example, the products that appear in a shopping transaction stream can be grouped using higher-level product categories. Although such explicit relationships have been exploited in previous works~\cite{yang2016revisiting} to improve the quality of the representations, they are not well-studied for the task of making representations compact. Also, for a given attribute of a multi-modal stream, we observe that a small fraction of units of the attribute appear during a short period of the stream compared to the total number of distinct units of the particular attribute. For example, the fraction of users appear in an updating window is less than 3\% of the total users for two multi-modal streams as shown in Figure~\ref{fig:online_frequency}. Likewise, the attributes show specific relationships and behaviours in multi-modal data streams. Our model is designed in a manner to \textbf{\textit{exploit such explicit/implicit relationships in multi-modal data streams.}}

\textbf{Contributions.} In this paper, we propose \textsc{METEOR}, a novel compact representation learning technique using multi-modal data streams, which:
\begin{itemize}
    \item learns compact online representations for multi-modal data units by sharing the parameters (i.e., basis vectors) inside the semantically meaningful groupings of the units. Also, \textsc{METEOR} is domain-agnostic, thus yielding consistent results with multi-modal streams from different domains;  
    \item proposes an architecture (\textsc{METEOR-Parallel}) to accelerate the learning of \textsc{METEOR}, which can learn embeddings of the same quality at twice the speed of \textsc{METEOR}; and
    \item can be easily extended to exploit explicit knowledge sources related to multi-modal data units by defining parameter sharing groups (implicitly defined if there is no such explicit knowledge source) based on the particular knowledge sources. Our results show that \textsc{METEOR} can further improve the quality of the compact embeddings by including explicit knowledge sources.  
\end{itemize}

\begin{figure*}[t]
    \centering
    \includegraphics[width=\linewidth]{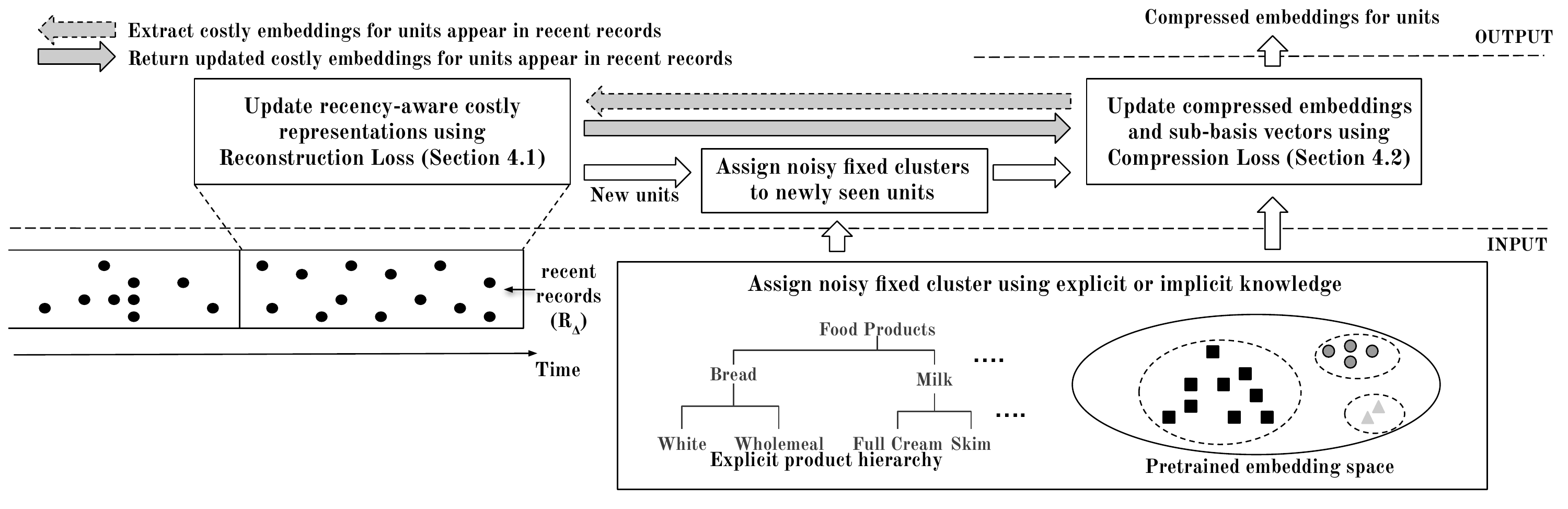}
    \caption{Overview of \textsc{METEOR}}
    \label{fig:overview}
    \vspace{-4mm}
\end{figure*}

\section{Related Work}\label{sec:related_works}
Although work on learning compressed coding systems began in the 1950's, such as error correction codes~\cite{hamming1950error}, and Hoffman codes~\cite{huffman1952method}, there has been recent progress in representation learning techniques~\cite{mikolov2013distributed, pennington2014glove} that encode each low-level unit with a continuous embedding vector. In such approaches, the number of parameters in the embeddings grows linearly with the number of low-level units. As a result, learning compact representations has become a popular research problem and recently been addressed by many previous works. Almost all the previous works on learning memory-efficient representations can be divided into two categories: (1) compositional embedding learning techniques; and (2) data compression techniques.

\textbf{(1) Compositional embedding learning techniques. }In these techniques, a set of basis vectors are learned, which are shared between all low-level units. Then the final embedding for a given unit is taken as a composition (e.g., linear combination) of the basis vectors. These techniques mainly differ from each other in the way the basis vectors are composed to generate final representations. \textsc{METEOR} also belongs to this category.

In some works~\cite{shi2019compositional,weinberger2009feature}, predefined hashing functions (e.g., division hashing and tree based hashing) are used to assign each low-level unit to one of the basis vectors. \textit{However, these approaches do not consider the semantics of the units when mapping to the basis vectors, thus they may blindly map vastly different units to the same basis vector.} This can lead to substantial loss of information and deterioration of the embeddings. To mitigate that, other recent works explore domain-specific sub-units with smaller vocabularies to define basis vectors. For example, in~\cite{svenstrup2017hash,kim2016character,sennrich2015neural}, the basis vectors are defined using characters and sub-words in a language. Then hash functions are defined to automatically
map words (or texts) to pre-defined bases, according to which the vectors are composed. However, such approaches may not be scalable for other domains as they use language-specific semantics when defining their compositions. In contrast, \textsc{METEOR} exploits the semantics of the low-level units in a domain-agnostic manner.

To learn task-independent compact embeddings, a few previous works~\cite{chen2018learning,shu2017compressing} attempt to jointly learn basis vectors and discrete codes for each unit, which defines the composition of the basis vectors. However, the main challenge of learning discrete codes to define composition is that they cannot be directly optimized via SGD like other parameters. In~\cite{chen2018learning,shu2017compressing}, the discrete encodings are relaxed using continuous relaxation techniques such as the Tempering-Softmax trick and Gumbel-Softmax trick. In~\cite{suzuki2016learning}, a similar technique has been proposed, which divides the aforementioned non-differentiable objective function into several solvable sub-problems, and sequentially solves each sub-problem. All these techniques require continuous costly embeddings (e.g., pretrained word embeddings) to be stored to guide their learning process. Also, we empirically observe that the latter is computationally expensive to learn in an online fashion. Thus, these techniques are not suitable for data streams, where the learning happens incrementally. In contrast, \textsc{METEOR} learns sparse continuous code vectors along with basis vectors, which could be trained in an online fashion. Also, to the best of our knowledge, \textsc{METEOR} is the first work on learning compact online embeddings using multi-modal data streams.

\textbf{(2) Data compression techniques. }In addition, some previous works~\cite{panahi2019word2ket,may2019downstream,acharya2019online} adopt data compression techniques (e.g., quantization, dimension reduction, sparse coding, quantum entanglement) to reduce the memory requirement to store embeddings. However, all these approaches linearly increase the size of the embedding table with the number of distinct units in the system, whereas \textsc{METEOR} introduces a small overhead with respect to the total number of distinct units.


\section{Problem Statement}

Let $R=\{r_1, r_2, r_3, ..., r_n, ...\}$ be a continuous stream of records that arrive in chronological order. Each record $r \in R$ is a tuple \linebreak$<a_0^r, a_1^r, ...., a_N^r>$, where $a_i^r$ is the $i^{th}$ attribute of $r$ and $N$ denotes the number of attributes of $r$.

Our problem is to learn embeddings for all possible units in each attribute, denoted as $A
_i (=\bigcup^{\forall r} \{a_i^r\})$ for the $i^{th}$ attribute, such that the embedding $v_x$ of a unit $x\in A_i$:
\begin{enumerate}
    \item is a $d$-dimensional vector ($d << \sum_{i=0}^N |A_i|$), where $|A_i|$ is the number of different units in $A_i$;
    \item preserves the co-occurrences of the attributes;
    \item is continuously updated as new records ($R_{\Delta}$) arrive to incorporate the latest information;
    \item is memory efficient with a memory complexity $ << O(d)$. 
\end{enumerate}

Consider a shopping transaction stream as an example for $R$, in which each record (i.e., transaction) $r$ can be characterized using a tuple consisting of three attributes $<t^r, u^r, p^r>$, where: (1) $t^r$ is the transaction time of $r$; (2) $u^r$ is the user id of $r$; and (3) $p^r$ is the set of products in the shopping basket of $r$. This work aims to jointly learn recency-aware vector representations for all possible units in each attribute (e.g., discretized\footnote{\label{foot1}Continuous attributes such as timestamps should be discretized to make them feasible for embedding} timestamps, products, and users in $R$) such that the co-occurring units have similar embeddings. 

\section{METEOR}
\textbf{Overview.} For a given attribute (e.g., products) in a multi-modal data stream (e.g., shopping transaction stream), \textsc{METEOR} groups all the possible low-level units (e.g., white bread and skim milk) of the particular attribute into a set of semantically meaningful clusters (i.e., \textit{noisy fixed clusters}). Then for each cluster, a set of basis vectors are trained to learn the \textit{costly embeddings} (i.e., conventional memory intensive embeddings) for the units in the particular cluster as a composition of the corresponding basis vectors. For each unit, the composition of the basis vectors (denoted as the \textit{compressed embedding}) is defined as a sparse continuous vector. 

Despite learning the compressed embeddings of the units directly, \textsc{METEOR} follows a computationally efficient sequential approach as shown in Figure~\ref{fig:overview}. For a given set of recent records $R_\Delta$, \textsc{METEOR} initially extracts the costly embeddings for the units that appeared in $R_\Delta$, then updates the extracted costly embeddings using $R_\Delta$ to incorporate the recent information. Subsequently, the recency-aware costly embeddings are used to update the corresponding compressed embeddings and basis vectors. This section discusses in detail how these steps are performed incrementally.

\subsection{Learning Costly Embeddings}\label{sec:costly_emb}
The desired embedding space of \textsc{METEOR} should be able to predict any attribute of a record given the other attributes (i.e., to preserve the first-order co-occurrences in records). Thus, the costly embeddings in \textsc{METEOR} are learned to recover the attributes of records as much as possible, which is formally elaborated as follows.

For a given record $r$, the embeddings are learned such that the units (e.g., product or user in a shopping transaction)  of $r$ can be recovered by looking at $r$'s other units. Formally, we model the likelihood for the task of recovering unit $x\in r$ given the other units $r_{-x}$ of $r$ as:
\begin{equation}
    p(x|r_{-x}) = \exp (s(x,r_{-x})/ \sum_{y\in X} \exp (s(y, r_{-x}))
\end{equation}
where $X$ is the type  (e.g., product or user in a shopping transaction) of $x$, and $s(x, r_{-x})$ is the similarity score between $x$ and $r_{-x}$. We define the $s(x, r_{-x})$ as $s(x, r_{-x})=v_x^Th_x$ where $h_x$ is mean embedding of $r$'s units except $x$.

Then, the final loss function for the attribute recovery task is the negative log likelihood of recovering all the attributes of the records in the current buffer $B$:
\begin{equation}\label{eq:pre_loss}
    O_{R_\Delta} = - \sum_{r \in R_\Delta} \sum_{x \in r} p(x|r_{-x})
\end{equation}

The objective function above is approximated using negative sampling (proposed in~\cite{mikolov2013distributed}) for efficient optimization using stochastic gradient descent (SGD). Then for a selected record $r$ and unit $x\in r$, the loss function is:
\begin{equation}\label{eq:loss}
    L_{recon} = - \log (\sigma (s(x, r_{-x}))) - \sum_{n\in N_x} \log (\sigma (-s(n, r_{-x})))
\end{equation}
where $\sigma(z) = \frac{1}{1 + \exp(-z)}$ and $N_x$ is the set of randomly selected negative units that have the type of $x$.

\textbf{Adaptive Optimization. }Since the loss function is incrementally optimized using a stream, only the recent records in the stream are used to update the embeddings. Hence, we adopt a novel adaptive strategy to optimize the loss function in  Equation~\ref{eq:loss} while alleviating overfitting to the recent records as follows.

For each record $r$, we compute the intra-agreement $\Psi_r$ of $r$'s attributes as:
 \begin{equation}
     \Psi_r = \frac{\sum_{x, y \in r, x\neq y} \sigma(v_{x}^{\top}v_{y})}{\sum_{x, y \in r, x\neq y} 1 }
 \end{equation}
 
 Then the adaptive learning rate of $r$ is calculated as,
 \begin{equation}
     lr_r = \exp (-\tau \Psi_r)* \eta
 \end{equation}
 where $\eta$ denotes the standard learning rate and $\tau$ controls the importance given to $\Psi_r$. If the representations have already overfitted to $r$, then $\Psi_r$ takes a higher value. Consequently, a low learning rate is assigned to $r$ to avoid overfitting. In addition, the learning rate for each unit $x$ in $r$ is further weighted using the approach proposed in AdaGrad~\cite{Duchi2010} to alleviate overfitting to frequent items. Then, the update for the $v_x$ at the $t^{th}$ timestep is:
 \begin{equation}
     v_x^{t+1} = v_x^t - \frac{lr_r}{\sqrt{\sum_{j=0}^{t-1} {(\frac{\partial L}{\partial v_x})}_j^2 + \epsilon}} {(\frac{\partial L}{\partial v_x})}_t
 \end{equation}
 
Our experimental results verify that the proposed optimization technique to accommodate online learning yields comparable (sometimes even superior) results compared to state-of-the-art sampling-based approaches\footnote{Sampling-based approaches require historical records to be stored, and the samples from the historical records are fed along with the recent records to update the embeddings incrementally.} without storing any historical records.

\subsection{Learning Compressed Embeddings}\label{sec:compress_emb}
The aforementioned framework learns an independent embedding vector for each unit, and is thus memory inefficient. To alleviate this problem, \textsc{METEOR} takes the following steps:
\begin{itemize}
    \item Step 1: Define a set of clusters (i.e., noisy fixed clusters) for a given attribute such that each unit of the particular attribute belongs to a single cluster.
    \item Step 2: Assign a set of basis vectors for each cluster.
    \item Step 3: Impose an additional constraint on costly embeddings of the units such that they are linear combinations of the basis vectors of the corresponding noisy fixed clusters.
\end{itemize}
Then, the compressed embedding $\hat{v}_x$ of a given unit $x$ is the set of weights related to the corresponding basis vectors (i.e., composition), which could be used to reconstruct the costly embedding $v_x$ given the basis vectors. The rest of this section discusses each step mentioned above in detail.

\subsubsection{Step 1 - Noisy Fixed Cluster Generation and Assignment} \label{subsubsec:step1}
For a given attribute $a$ (e.g., products in a shopping transaction stream), \textsc{METEOR} defines a set of clusters $C^a = \{C_1^a,C_2^a,....C_{|C|}^a\}$ and a cluster assignment function $g^a:x\rightarrow c^x$ (where $c^x \in C^a$), which assigns each unit $x$ (e.g., white bread or skim milk if the attribute is a product in a shopping transaction stream) in $a$ to a single cluster in $C^a$. $C^a$ and $g^a$ can be determined using either (1) explicitly available domain knowledge or (2) implicitly using a pretrained embedding space.

\textbf{(1) Explicit clusters. }These are the clusters generated using an explicitly available grouping scheme. As an example, products (i.e., a modality of a shopping transaction stream) can be grouped using explicitly available product categories. Both $C^a$ and $g^a$ can be determined using such a grouping, which do not generally change over time.

\textbf{(2) Implicit clusters. }The clusters in $C^a$ can be generated using a clustering algorithm like KMeans~\cite{lloyd1982least} by clustering a pretrained embedding space (by optimizing Equation~\ref{eq:loss} using a subset $R_{pre}$ of $R$) of $a$. Then $g^a$ is defined such that it assigns each unit in the embedding space to the closest cluster based on the Euclidean distance to each cluster. 

\textbf{Assumption 1.} \textsc{METEOR} assumes that the units do not change their clusters once they have been assigned to a noisy fixed cluster. This assumption is valid for explicit clusters. However, it may not be true for implicit clusters, as they are generated using the embeddings of the units, which are updated over time. As shown by~\cite{guo2017improved}, allowing hard cluster assignments to change over time could degrade the embedding space due to the sudden changes of the embeddings, which are incrementally learned over time. To preserve the validity of this assumption for the implicit clusters, we set the number of noisy fixed clusters to a small number (i.e., $< 1\%$ of the total number of units) to make the size of clusters big enough to minimize cluster changes over time. We empirically observe that Assumption 1 restricts only a few changes in the cluster assignments under the aforementioned setting, as shown in Figure~\ref{fig:cluster_changes}. In Figure~\ref{fig:cluster_changes}, the users in the TF dataset are clustered into 64 clusters based on their embeddings using KMeans  (with the initialization of cluster centres using the cluster centres of the previous updating window) at the end of each updating window,  and we observe that at most $0.06\%$ users change their clusters over time.


\begin{figure}[t]
\begin{tikzpicture}
\begin{axis}[
    width=6.8cm,
    height=4cm,
    ylabel={$\#\;of\;cluster\;changes$},
    xlabel={$updating\;window\;id$},
    xtick={0,10,20,30,40,50},
    ymajorgrids=true,
    grid style=dashed,
]

\addplot[mark size=1pt] table [col sep=space] {cluster_changes.dat};
\end{axis}
\end{tikzpicture}
\caption{A smaller fraction of users in the TF dataset, which consists of $\bf9238$ users in total, change their clusters over different updating windows ($\bf\Delta W=1\;day$) if cluster changes are allowed (without Assumption 1)}\label{fig:cluster_changes}
\vspace{-3mm}
\end{figure}
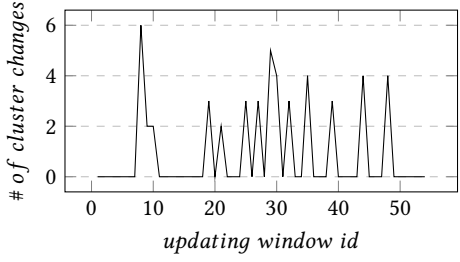

\begin{figure}[t]
\begin{tikzpicture}
\begin{axis}[
    width=7cm,
    height=4cm,
    xlabel={$i\; (cluster\;id)$},
    ylabel={$|C_i|\;(cluster\;size)$},
    ymajorgrids=true,
    grid style=dashed,
]

\addplot[mark size=1pt] table [col sep=space] {cluster_size.dat};
\end{axis}
\end{tikzpicture}
\caption{The sizes of the noisy fixed clusters (64 clusters) generated for users in TF Dataset are uneven}\label{fig:unbalanced_clus}
\vspace{-4mm}
\end{figure}
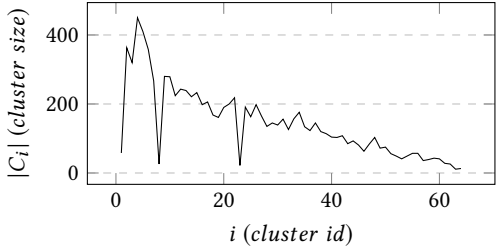

\subsubsection{Step2 - Basis Vector Assignment}\label{subsubsec:step2}For each noisy fixed cluster $C_i^a$ of attribute $a$, a set of $d$-dimensional vectors $B_i^a$ are assigned to represent $C_i^a$'s  basis. Let $|B_i^a|$ be the number of basis vectors in $B_i^a$, then $B_i^a \in R^{d\times|B_i^a|}$ (where $d>>|B_i^a|$).    
    
\textbf{Number of basis vectors per cluster. }We empirically observe that the noisy fixed clusters generated using either explicit or implicit approaches are uneven in size, as shown in Figure~\ref{fig:unbalanced_clus}. Hence, when assigning the basis vectors for different noisy fixed clusters, \textsc{METEOR} assigns more basis vectors to the large clusters using Equation~\ref{eq:n_basis_vecs} (based on the cluster sizes at the initialization) to accommodate uniform encoding for all the units. Let $K^a$ be the total number of basis vectors assigned to different noisy fixed clusters of attribute $a$, then the number of basis vectors for $C^a_i$, $|B_i^a|$, is calculated as,
\begin{equation}\label{eq:n_basis_vecs}
    |B_i^a| = ceil(K^a*\frac{|C_i^a|}{\sum_{\forall j} |C_j^a|})
\end{equation}
where $ceil(.)$ is the standard ceiling operation.

\begin{algorithm}[t]
 \LinesNumbered
 \SetAlgoLined
 \SetKwInput{Input}{Input}
 \SetKwInput{Output}{Output}
 \Input{The noisy fixed cluster assignments $C$\linebreak The current compressed embeddings $\hat{V}$\linebreak
        The current basis vectors $B$
        \linebreak A collection of new records $R_\Delta$
      }
 \Output{The updated $\hat{V}$ and $B$}
 $V \leftarrow  \emptyset$;\\
 \For{$unit\;x$ in $R_\Delta$}{
        
        $v_x \leftarrow$  
        $\left\{\begin{array}{ll}
        initialize \; randomly & \quad \text{if }x\text{ is a new unit}\\
        compute\;using\; Eq.~\ref{eq:mem_emb}  & \quad \text{otherwise}\\
        \end{array}\right.$
        \\
        $V \leftarrow \{v_x\} \cup V$
 }
 \tcp{Optimize $L_{recon}$ using $R_{\Delta}$}
 \For{$epoch \; from \; 1 \; to \; N$}{
      \For{$r$ in $R_\Delta$}{
      Update $V$ to recover $r$'s attribute using Eq.~\ref{eq:loss};\\ 
      }
  }
  \tcp{Assign noisy fixed clusters for new units}
  \For{$v_x \in V$}{
      \If{$x$ is a new unit}{
      $Assign\;v_x\;to \; the\; closest\; noisy\;fixed\;cluster$\\ 
      }
  }
  \tcp{Optimize $L_{comp}$ using $V$}
 \For{$v_x \in V$}{
      Update $\hat{V}$ and $B$ using Eq.~\ref{eq:compress};\\ 
  }
 Delete V;\\
 Return $\hat{V}\; and\; B$
 \caption{METEOR Learning}\label{algo:algo1}
\end{algorithm}

\subsubsection{Step 3- Learning Compressed Embeddings and Basis Vectors}\label{subsubsec:step3}
For each unit $x \in C_i^a$, \textsc{METEOR} learns $x$'s costly embedding $v_x$ as a linear combination of $B_i^a$,
\begin{equation}\label{eq:mem_emb}
    v_x=B_i^a\cdot \hat{v}_x
\end{equation}
where $\hat{v}_x$ is the memory-efficient compressed embedding of $x$, $\hat{v}_x \in R^{|B_i|}$ and $|B_i| << d$. A trivial way to learn the memory-efficient embeddings $\hat{V}$ and basis vectors $B$ is by replacing the embeddings in Equation~\ref{eq:loss} using Equation~\ref{eq:mem_emb}. However, that approach is computationally expensive to perform in an online fashion as it introduces many matrix multiplications into the loss function.

Thus, \textsc{METEOR} decomposes the original loss function as follows, 
\begin{equation}\label{eq:final_loss}
    L = L_{recon} + L_{comp}
\end{equation}
where $L_{comp}$ for a given unit $x$ of attribute type $a$ is defined as,
\begin{equation}\label{eq:compress}
    L_{comp} = (B_i^a \cdot \hat{v}_x - v_x)^2 + \lambda*||\hat{v}_x||
\end{equation}
where $\lambda$ is the weight given to the L1 regularization term. \textsc{METEOR} imposes L1 (i.e., Lasso) regularization on $\hat{V}$ to make the memory-efficient representations $\hat{V}$ as sparse as possible. This yields memory-efficient embeddings with a small fraction of non-zero values, which further reduces the memory requirement to store $\hat{V}$ using sparse matrix storage formats. 

\textbf{Initialization of Compressed Embeddings and Basis Vectors. }METEOR exploits the costly embedding in a pretrained embedding space,\footnote{The pretraining can be performed in a large server with sufficient memory capacity, as it needs to be performed only once.} which is used to generate noisy fixed clusters, and iteratively optimizes $L_{comp}$ in Equation~\ref{eq:compress} using Adagrad to initialize $\hat{V}$ and $B$.

\textbf{Incremental Learning of Compressed Embeddings and Basis vectors.} Then the compressed embeddings and the basis vectors are learned incrementally using the newly arrived records from the stream to incorporate recent information into the embeddings. For a given new set of records $R_\Delta$, \textsc{METEOR} adopts a sequential approach as shown in Algorithm~\ref{algo:algo1} to learn the model, instead of jointly optimizing $L$ in Equation~\ref{eq:final_loss}. Initially, \textsc{METEOR} produces the costly embeddings for the units (using Equation~\ref{eq:mem_emb}) that appear in $R_\Delta$ using the compressed embeddings and basis vectors returned at the end of the previous updating window (Line 2-5 in Algo~\ref{algo:algo1}). For new units, \textsc{METEOR} randomly initializes their costly embeddings. Then the costly embeddings are updated using Equation~\ref{eq:loss} to reconstruct the recent records. Then the new units of attribute type $a$ are assigned to the noisy fixed clusters using the cluster assignment function $g^a$, introduced in Section~\ref{subsubsec:step1} (Line 11-14). Subsequently, $\hat{V}$ and $B$ are updated using the recently updated costly embeddings by minimizing the compression loss in Equation~\ref{eq:compress} (Line 16-18). 

Within an updating window, \textsc{METEOR} maintains costly embeddings only for the units appearing within the particular window, which is a small fraction compared to the total number of units (see Figure~\ref{fig:online_frequency}). At the end of the updating window, \textsc{METEOR} deletes the costly embeddings and only retains the compressed embeddings and basis vectors in memory, which requires considerably less memory (discussed in detail in Section~\ref{subsec:complexity}). Thus, \textsc{METEOR} can be deployed on low-memory devices to learn recency-aware representations for different domains.  

\begin{figure}[t]
    \centering
    \includegraphics[width=0.85\linewidth, height=6cm]{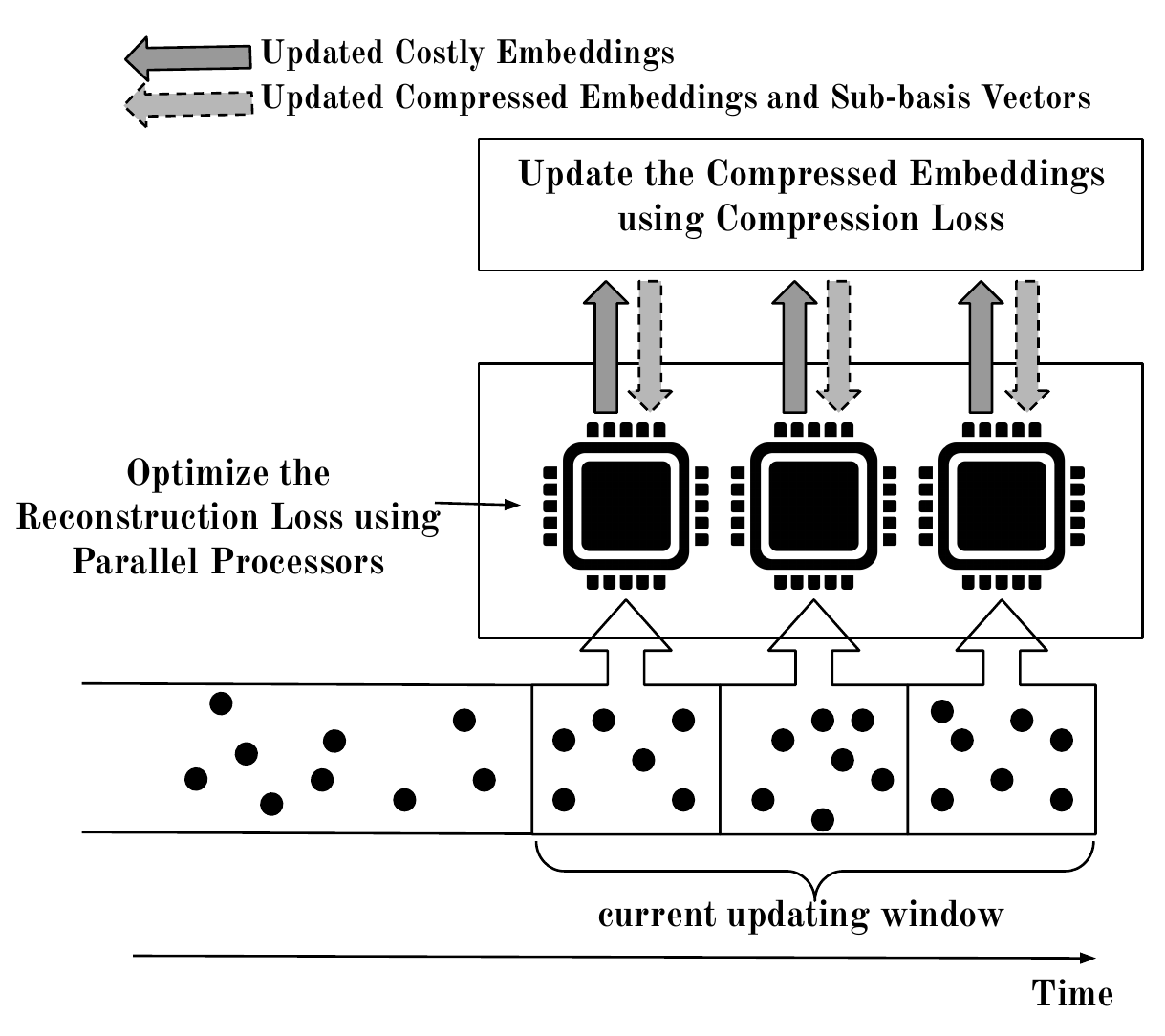}
    \vspace{-2mm}
    \caption{Architecture of \textsc{METEOR-Parallel}}
    \label{fig:meteor_parallel}
    \vspace{-4mm}
\end{figure}

\subsection{\textsc{METEOR-Parallel}}
The other main challenge for a system like \textsc{METEOR} is scalability to high-speed data streams. The number of records processed by a system like \textsc{METEOR} should be at least similar to the rate of the stream to accommodate all the records in the stream. Since the rates of multi-modal streams can change significantly with the domain, a domain-agnostic model like \textsc{METEOR} should be able to adapt to a wide range of data rates. Thus, this section proposes a way to accelerate the learning process of \textsc{METEOR} using a parallel processing architecture with different memory domains. We denote this version of \textsc{METEOR} as "\textsc{METEOR-Parallel}" for the rest of this paper\footnote{The architecture of \textsc{METEOR-Parallel} is different from the conventional multi-threading architectures, which mostly use shared memory. Inside of a process in \textsc{METEOR-Parallel}, conventional multi-threading can be still applied for further acceleration. However, we emulate the architecture of  \textsc{METEOR-Parallel} using multiple threads in this paper. We leave the performance of \textsc{METEOR-Parallel} on a real computer cluster as future work.}. 

As presented in the above section, \textsc{METEOR} optimizes the decomposed objective function in Equation~\ref{eq:final_loss} using two sequential steps: (1) update the costly embeddings to reconstruct the records in the current updating window; and (2) optimize the compression loss to update the compressed embeddings and the basis vectors. We have empirically identified that the latter step has a considerably lower time complexity compared to the former. Thus the bottleneck of the \textsc{METEOR} learning process lies with the former step. Also, we observed that learning Step 1 using parallel processors with separate memory units (e.g., a cluster of computers) does not deteriorate the quality of the compressed embeddings, and preserve the performance for downstream applications (see Figure~\ref{fig:results_meteor-parallel}). Hence, the architecture of \textsc{METEOR-Parallel} consists of multiple parallel processes to perform the first step and a single processor to perform the latter as shown in Figure~\ref{fig:meteor_parallel}.  

Then for a given set of records in $R_\Delta$, \textsc{METEOR-Parallel} divides the records into $p$ number of parallel processes. Each processor reads the most recent compressed embeddings and basis vectors to their own memory. Then each processor updates the costly embeddings (generated from the compressed embeddings and the basis vectors) to reconstruct the records arriving to each processor. Subsequently, all the updated costly embeddings are passed to the next stage, which centrally optimizes the compression loss to update the compressed embeddings and the basis vectors.

\subsection{Complexity of \textsc{METEOR}}\label{subsec:complexity}
In this section, we analyse the complexity of \textsc{METEOR} and \textsc{METEOR-Parallel} using a multi-modal stream, in which each attribute $a$ has $|A^a|$ distinct units. 

\subsubsection{Memory Complexity}
Assume that the number of noisy fixed clusters is $|C^{a}| (\approx 1\%\times|A^a|)$ and the total number of basis vectors shared between noisy fixed clusters of attribute $a$ is $K^{a}$. The space complexity to store conventional costly embeddings is $O(d\times(\sum_{\forall a} |A^a|))$. In contrast, the space complexity of \textsc{METEOR} is \linebreak$O(\sum_{\forall a} K^a (|A^a|/|C^a| + d))$ in the average case, where $K^a/|C^a| << d$ and $K^a << |A^a|$. The memory complexity per processor in \textsc{METEOR-Parallel} remains the same.

\subsubsection{Time Complexity}The time complexity of \textsc{METEOR} consists of: (1) $O(ENM^2d \max(R_\Delta))$ to optimize the reconstruction loss (Step 1); and (2) $O(Ed \sum_{\forall a}\max(|a_\Delta|) K^a/|C^a|)$ to optimize the compression loss (Step 2), where $E, N,$ and  $M$ are the number of epochs, the number of negative samples, and the maximum number of attributes of a record, respectively. $|a_\Delta|$ is the distinct units of attribute type $a$ that appear in $R_\Delta$, which is a smaller fraction of $|A^a|$ as shown in Figure~\ref{fig:online_frequency}. The complexity of \textsc{METEOR-Parallel} with $p$ processes remains same as for Step 2, and $O(ENM^2d \max(R_\Delta)/p)$ for Step 1.

\section{Experimental Setup}\label{sec:experiments}

\textbf{Datasets.} We conduct our experiments using three shopping transaction datasets: 

\begin{itemize}
\item Complete Journey Dataset\footnote{\href{https://www.dunnhumby.com/careers/engineering/sourcefiles}{https://www.dunnhumby.com/careers/engineering/sourcefiles}} (CJ) contains transactions at a retailer by 2,500 frequent shoppers over two years.
\item Ta-Feng Dataset\footnote{\href{http://www.bigdatalab.ac.cn/benchmark/bm/dd?data=Ta-Feng}{http://www.bigdatalab.ac.cn/benchmark/bm/dd?data=Ta-Feng}} (TF) includes shopping transactions of the Ta-Feng supermarket from November 2000 to February 2001.
\item InstaCart Dataset\footnote{\href{https://www.instacart.com/datasets/grocery-shopping-2017}{https://www.instacart.com/datasets/grocery-shopping-2017}} (IC) contains the shopping transactions of Instacart, an online grocery shopping centre, in 2017.
\end{itemize}
and three geo-tagged Twitter datasets collected from three urbanized cities:
\begin{itemize}
\item LA Dataset\footnote{\label{la_link}\href{https://drive.google.com/file/d/0Byrzhr4bOatCRHdmRVZ1YVZqSzA/view}{https://drive.google.com/file/d/0Byrzhr4bOatCRHdmRVZ1YVZqSzA/view}} (LA) contains around 1.2 million geo-tagged tweets from Los Angeles during the last quarter of 2014.
\item NY Dataset\footref{la_link} (NY) includes 1.5 million geo-tagged tweets collected from New York during the last quarter of 2014. 
\item MB Dataset\footref{la_link} (MB) includes 263,363 geo-tagged tweets collected from Melbourne during the period from 2016 November to 2018 January.
\end{itemize}

The descriptive statistics of the datasets are shown in Table~\ref{tab:dataset_statistics}. The evaluation based on the datasets from different domains supports the domain-agnostic behaviour of the proposed model. As can be seen, the datasets from the same domain also have significantly different statistics. For example, TF has a shorter collection period and IC has a larger user base. This helps to evaluate the performance of \textsc{METEOR} in different environment settings.

\begin{table}[t]
\scriptsize
\centering
\caption{Descriptive statistics of the datasets}\label{tab:dataset_statistics} 
\vspace{-2mm}
\begin{tabular}{l|c|c|c|c}
\hline
\bf Shopping Transaction Datasets & \# Transactions & \# Users &   \# Items & \# Baskets \\
\hline
\hline
Complete Journey (CJ) & 2,595,733& 2,500& 92,339  & 276,483\\
\hline
Ta-Feng (TF) & 464,118 & 9,238 & 7,973 & 77,202\\
\hline
Instacart (IC) &33,819,306 &206,209 & 49688 & 3,421,083\\
\hline
\hline
\bf Geo-tagged Twitter Datasets & \# Tweets  & \# Users &   \# Locations& \# Words \\
\hline
Los Angeles (LA) & 1,188,405 & 153,626&18,271 &353,586\\
\hline
New York (NY) & 1,500,000& 176,597& 19,108&419,923\\
\hline
Melbourne (MB) & 263,363& 27,056& 1,101&98,698\\
\hline
\end{tabular}
\vspace{-4mm}
\end{table}

\textbf{Baselines. }We compare \textsc{METEOR} with the following methods:
\begin{itemize}
    \item \textsc{Dim\_Reduct} reduces the size of the costly embedding vectors ($d$) and learns the embeddings as shown in Section~\ref{sec:costly_emb}.
    \item \textsc{$m$-bit Quantization} evenly quantizes the values in the costly embeddings into $2^m$-bins as post processing.
    \item \textsc{Hash Trick} adopts modulo-division hashing to assign low-level units to clusters and a shared $d-$dimensional embedding is trained for each cluster. The divisor of the hashing function for an attribute $a$ is set as $\gamma\times A$, where $A$ is the total number of distinct units in $a$, and $\gamma$ defines the value of the divisor as a proportion of $A$.
    \item \textsc{DCN+Hard Clustering} adopts the deep clustering approach proposed in~\cite{yang2017towards} to learn the costly embeddings. At the end of each updating window, the embedding space related to an attribute $a$ is clustered into $\gamma\times A$ clusters and the embeddings of the units are replaced by the corresponding cluster centers.
\end{itemize}
We compare a few online learning techniques with the proposed costly embedding learning approach (\textsc{METEOR-Full}) in \textsc{METEOR}:
\begin{itemize}
    \item \textsc{METEOR-Decay} and \textsc{METEOR-Info} adopt SGD optimization with the sampling-based online learning methods proposed in~\cite{zhang2017react} and~\cite{silva2019fullustar}, respectively.
    \item \textsc{METEOR-Cons} adopts SGD optimization with the constraint-based online learning approach proposed in~\cite{zhang2017react}.
\end{itemize}

\begin{table*}[]
\footnotesize
    \centering
    \caption{Results for \textit{intra-basket item retrieval}}
    \vspace{-2mm}
    \begin{tabular}{|cc|c|c|c|c|c|c|c|c|c|c|}
    \cline{4-12}
    \multicolumn{3}{c|}{}&\multicolumn{3}{c|}{CJ Dataset}&\multicolumn{3}{c|}{TF Dataset}&\multicolumn{3}{c|}{IC Dataset}\\
    \hline
    method& \begin{tabular}[c]{@{}c@{}}parameter(s) for \\ reducing model size\end{tabular}& \begin{tabular}[c]{@{}c@{}}Additional \\memory for \\a\ new unit\end{tabular}& \begin{tabular}[c]{@{}c@{}}Model\\size\end{tabular}&MRR&R@1& \begin{tabular}[c]{@{}c@{}}Model\\size\end{tabular}&MRR&R@1& \begin{tabular}[c]{@{}c@{}}Model\\size\end{tabular}&MRR&R@1\\
    \hline
    \hline
    \textsc{METEOR-Full} &costly embeddings (d=300)&\bf2.34KB& \bf217MB&\bf0.6013&\bf0.4325&\bf39MB&\bf0.5166&\bf0.3466&\bf586MB&\bf0.7478&\bf0.5859\\
    \textsc{METEOR-Info} &costly embeddings (d=300)&2.34KB& 217MB&0.5991&0.4275&39MB&0.4046&0.2205&586MB&0.7482&0.5852\\
    \textsc{METEOR-Decay} &costly embeddings (d=300)&2.34KB& 217MB&0.5984&0.4221&39MB&0.402&0.2186&586MB&0.7117&0.5442\\
    \textsc{METEOR-Cons} &costly embeddings (d=300)&2.34KB& 217MB&0.4610&0.2742&39MB&0.3996&0.2031&586MB&0.5942&0.4193\\
    \hline\hline
    \textsc{METEOR} & $|C^a|=0.5\%|A^a|$, $K^a=10\%|A^a|$ &\bf0.16KB &\bf78MB&\bf0.5974&\bf0.4293&\bf14MB&\bf0.5013&\bf0.3287&\bf112MB&\bf0.7203&\bf0.5507\\
     & $|C^a|=1\%|A^a|$, $K^a=10\%|A^a|$ &\bf0.08KB& \bf48MB&\bf0.587&\bf0.4098&\bf9MB&\bf0.4961&\bf0.3203&\bf71MB&\bf0.7084&\bf0.5411\\
     \hline\hline
     \textsc{Dim\_Reduct}
     & d = 100&0.76KB&72MB&0.5773&0.4044&13MB&0.4762&0.2940&195MB&0.6923&0.5213\\
     & d = 50& 0.38KB&36MB&0.5495&0.3664&7MB&0.4509&0.2713&98MB&0.6628&0.4825\\
     & d = 25& 0.19KB&18MB&0.5178&0.3323&4MB&0.4228&0.2305&49MB&0.6284&0.4381\\
     \hline
     \textsc{Quantization}
     & 8 bit quant.&0.59KB&54MB&0.5677&0.3722&10MB&0.4672&0.2892&147MB&0.6836&0.5077\\
     & 4 bit quant.&0.29KB&27MB&0.5453&0.3502&5MB&0.4478&0.2700&73MB&0.6447&0.4603\\
     & 2 bit quant.&0.15KB&14MB&0.4321&0.2801&3MB&0.3217&0.1413&37MB&0.5573&0.3705\\
    \hline
    \textsc{Hash Trick}
    & $\gamma= 30\%$ &4B& 65MB&0.5376&0.3487&12MB&0.4335&0.2535&176MB&0.6558&0.4727\\
    & $\gamma= 20\%$ &4B& 43MB&0.4993&0.3106&8MB&0.3711&0.1987&117MB&0.6248&0.4376\\
    & $\gamma= 10\%$ &4B& 22MB&0.4677&0.2988&4MB&0.3417&0.1786&58.6MB&0.5688&0.3848\\
    \hline
    \textsc{DCN + Hard Clustering}
    & $\gamma= 30\%$ &4B& 65MB&0.5477&0.3588&12MB&0.4577&0.2724&176MB&0.6731&0.4906\\
    & $\gamma= 20\%$ &4B& 43MB&0.5321&0.3411&8MB&0.4122&0.2236&117MB&0.6482&0.4583\\
    & $\gamma= 10\%$ &4B& 22MB&0.4882&0.3075&4MB&0.3876&0.2033&58.6MB&0.5883&0.4017\\
    \hline
    \end{tabular}
    \label{tab:intra-basket}
    \vspace{-2mm}
\end{table*}

\textbf{Parameter Settings. }The two main parameters of \textsc{METEOR} are $|C^a|$ and $K^a$. For most of the experiments, we set $|C^a|=1\%|A^a|$ and $K^a=10\%*|A^a|$ ($|A^a|$ is the total distinct units of attribute $a$ in the pretrained embedding space), otherwise the parameter values are specified. We present the results with different $|C^a|$ and $K^a$ values in Figure~\ref{fig:results_hyperparameter}. The weight given to the sparsity constraint $\lambda$ is set to 0.001 in \textsc{METEOR} after performing a grid search over $[0, 0.0001, 0.001, 0.01, 0.1, 1]$ values. 

In addition, all the aforementioned techniques share three common parameters (default values are given in brackets): (1) the costly
embedding dimension $d$ (300), (2) the SGD learning rate $\eta$ (0.05), (3) the negative samples $|N_x|$ (3), and (4) the number of epochs
$N$ (50). We set $\tau = 0.1$  for sampling based online learning techniques. 

\textbf{Evaluation Metrics. } \textsc{METEOR} is quantitatively evaluated using two retrieval tasks: (1) \textit{intra-basket item retrieval task} for shopping transaction datasets; and (2) \textit{location retrieval task} for geo-tagged Twitter datasets. Both tasks follow a similar experimental setup. Similar to previous work~\cite{zhang2017react}, we adopt the following procedure to evaluate the performance of each retrieval task. For each record in the test set, we select one unit (e.g., a product for \textit{intra-basket item retrieval} or the location for \textit{location retrieval}) as the target prediction and the rest of the units of the record as the context. We mix the ground truth target unit with a set of $M$ negative samples (i.e., a set of units that have the type of the ground truth) to generate a candidate pool to rank. $M$ is set to 10 for all the experiments. Then the size-$(M+1)$ candidate pool is sorted to get the rank of the ground truth. The average similarity of each candidate unit to the context of the corresponding test instance is used to produce the ranking of the candidate pool. Cosine similarity is used as the similarity measure for all the baselines.

If the model is well trained, then higher ranked units are most likely to be the ground truth. Hence, we use two different evaluation metrics to analyze the ranking performance: 
\begin{enumerate}
    \item Mean Reciprocal Rank (MRR) $ =\frac{\sum_{q = 1}^Q 1/rank_i}{|Q|}$
    \item Recall@k (R@k) $ =\frac{\sum_{q = 1}^Q min(1, \lfloor k/rank_i\rfloor)}{|Q|}$
\end{enumerate}
where $Q$ is the set of test queries and $rank_i$ refers the rank of the ground truth label for the $i$-th query. $\lfloor.\rfloor$ is the floor operation. A good ranking performance should yield higher values for the both evaluation metrics. 

We divide the records in each data stream into different updating windows such that each window has a length of $\Delta w$ ($\Delta w= 1\; day$ for shopping transaction datasets and $\Delta w= 1\; hour$ for geo-tagged Twitter datasets).
The first half of the period for each dataset is used to pretrain costly embeddings, which are subsequently used to produce noisy-fixed clusters. We randomly select 20 query updating windows from the second half of the period for each dataset, and all the records in the randomly selected time windows are used as test instances. For each query window, we only use the records that arrive before the query window to train different models. We ignore timestamps in both types of streams in the embedding as they do not substantially affect the performance of the tasks. The locations in geo-tagged Twitter streams are discretized into $300m\times300m$ small grids to make them feasible for embedding.
\vspace{-1mm}
\section{Results}

\begin{table*}[t]
\footnotesize
    \centering
    \caption{Results for \textit{location retrieval}}
    \vspace{-2mm}
    \begin{tabular}{|cc|c|c|c|c|c|c|c|c|c|c|}
    \cline{4-12}
    \multicolumn{3}{c|}{}&\multicolumn{3}{c|}{LA Dataset}&\multicolumn{3}{c|}{NY Dataset}&\multicolumn{3}{c|}{MB Dataset}\\
    \hline
    method& \begin{tabular}[c]{@{}c@{}}parameter(s) for \\ reducing model size\end{tabular}& \begin{tabular}[c]{@{}c@{}}Additional \\memory for \\a\ new unit\end{tabular}& \begin{tabular}[c]{@{}c@{}}Model\\size\end{tabular}&MRR&R@1& \begin{tabular}[c]{@{}c@{}}Model\\size\end{tabular}&MRR&R@1& \begin{tabular}[c]{@{}c@{}}Model\\size\end{tabular}&MRR&R@1\\
    \hline
    \textsc{METEOR-Full} &costly embeddings (d=300)&\bf2.34KB& \bf1118MB&\bf 0.7997&\bf0.6943& \bf1313MB&\bf0.7956&\bf0.6894& \bf274MB& \bf0.6157&\bf0.4476 \\
    \hline
    \textsc{METEOR-Info} &costly embeddings (d=300)&2.34KB& 1118MB& 0.8053& 0.6943& 1313MB& 0.8027& 0.6927& 274MB&0.6021 & 0.4388\\
    \hline
    \textsc{METEOR-Decay} &costly embeddings (d=300)&2.34KB& 1118MB& 0.7614& 0.6488& 1313MB& 0.775& 0.6590& 274MB&0.5808 & 0.4169\\
    \hline
    \textsc{METEOR-Cons} &costly embeddings (d=300)&2.34KB& 1118MB& 0.7602& 0.6488& 1313MB& 0.7701& 0.6562& 274MB&0.5762 & 0.4001\\
    \hline\hline
    \textsc{METEOR} & $|C^a|=0.5\%|A^a|$, $K^a=10\%|A^a|$ &\bf 0.16KB & \bf 204MB& \bf0.7899& \bf0.6858& \bf241MB& \bf0.7943& \bf0.6805&\bf62MB& \bf0.6004& \bf0.4366\\
     & $|C^a|=1\%|A^a|$, $K^a=10\%|A^a|$ &\bf0.08KB& \bf128MB& \bf0.7701& \bf0.6631&\bf152MB & \bf0.7803& \bf0.6714&\bf31MB& \bf0.5897 &\bf0.4287\\
     \hline\hline
     \textsc{Dim\_Reduct}
     & d = 100&0.76KB& 373MB& 0.7426& 0.6134& 438MB& 0.7398& 0.6188& 91MB& 0.5403&0.3614\\
     & d = 50& 0.38KB& 187MB& 0.7116& 0.5683& 219MB& 0.7073&0.5692 & 46MB& 0.5022&0.3245\\
     & d = 25& 0.19KB& 93MB& 0.6882& 0.5385& 109MB&0.6702 &0.5173 & 23MB& 0.4595&0.2806\\
     \hline
     \textsc{Quantization}
     & 8 bit quant.&0.59KB& 280MB& 0.7262& 0.5882&328MB & 0.7288& 0.5946& 69MB& 0.5333&0.3572\\
     & 4 bit quant.&0.29KB& 140MB& 0.6903& 0.5483&164MB& 0.6929& 0.5493& 34MB& 0.4859&0.3006\\
     & 2 bit quant.&0.15KB&  70MB& 0.5883& 0.4277&82MB &0.5892 &0.4196& 17MB& 0.4007&0.2177\\
    \hline
    \textsc{Hash Trick}
    & $\gamma= 30\%$ &4B& 335MB& 0.7001&0.5481 & 394MB& 0.7065& 0.5582& 82MB& 0.5338&0.3520\\
    & $\gamma= 20\%$ &4B& 224MB& 0.6638& 0.5083& 263MB& 0.6690&0.5173& 55MB& 0.4961&0.3169\\
    & $\gamma= 10\%$ &4B& 112MB& 0.6104& 0.4854& 131MB& 0.6152& 0.4502&27MB& 0.4517&0.2752\\
    \hline
    \textsc{DCN + Hard Clustering}
    & $\gamma= 30\%$ &4B& 335MB& 0.7177& 0.5764& 394MB& 0.7208& 0.5872& 82MB& 0.5382&0.3520\\
    & $\gamma= 20\%$ &4B& 224MB&  0.6843& 0.5375& 263MB& 0.6943& 0.5493& 55MB& 0.5099&0.3287\\
    & $\gamma= 10\%$ &4B& 112MB& 0.6429& 0.5104&131MB&0.6544 & 0.4921& 27MB& 0.4728&0.2970\\
    \hline
    \end{tabular}
    \label{tab:location}
    \vspace{-2mm}
\end{table*}

\begin{table}[]
\footnotesize
    \centering
    \caption{Results for \textit{intra-basket item retrieval} with explicit product clusters (categories 92,339 products in to 2384 product categories) in CJ Dataset as noisy fixed clusters}
    \vspace{-2mm}
    \begin{tabular}{|c|c|c|c|c|}
    \hline
    \begin{tabular}[c]{@{}c@{}}The variant of\\ \textsc{METEOR}\end{tabular}&  \begin{tabular}[c]{@{}c@{}}Average \\memory for\\ a new unit\end{tabular}&\begin{tabular}[c]{@{}c@{}}Model\\size\end{tabular}&MRR&R@1\\
    \hline
    Using costly embeddings (d=300)&2.34KB& 217MB&0.6013&0.4325\\
    \hline
    Without explicit product clusters &&&&\\
     $|C^a|=0.5\%|A^a|$, $K^a=10\%|A^a|$ &0.25KB& 78MB&0.5974&0.4293\\
     $|C^a|=1\%|A^a|$, $K^a=10\%|A^a|$ &0.13KB& 48MB&0.587&0.4098\\
    \hline
    With explicit product clusters &&&&\\
     $|C^a|=0.5\%|A^a|$, $K^a=10\%|A^a|$ &0.25KB& 59MB&\bf0.6097&\bf0.4401\\
     $|C^a|=1\%|A^a|$, $K^a=10\%|A^a|$ &0.13KB& 37MB&0.6028&0.4337\\
    \hline
    \end{tabular}
    \label{tab:explicit_clusters}
    \vspace{-2mm}
\end{table}

\begin{table}[]
\footnotesize
    \centering
    \caption{Ablation Study of Elements in METEOR}
    \vspace{-2mm}
    \begin{tabular}{|l|c|c|c|c|}
    \cline{2-5}
    \multicolumn{1}{c}{}&\multicolumn{2}{|c|}{CJ Dataset}&\multicolumn{2}{|c|}{LA Dataset}\\
    \hline
    Method& MRR&R@1& MRR&R@1\\
    \hline
      \textsc{METEOR}  & 0.587&0.4098 &0.7701&0.6631\\
    \hline
     (-) weighted basis vector assignment &0.5703&0.3995&0.7577&0.6304\\
     (-) sparsity constraint &0.5852&0.4098&0.7695&0.6631\\
    \hline
    \end{tabular}
    \label{tab:ablations_study}
    \vspace{-4mm}
\end{table}

\textbf{(1) Compressed Embedding Learning in \textsc{METEOR}. } Table~\ref{tab:intra-basket} and Table~\ref{tab:location} show the results collected for \textit{intra-basket item retrieval} and \textit{location retrieval} respectively. \textsc{METEOR} shows significantly better results than the comparable baseline models, which are the models with similar size. For example, if we consider the results collected for \textit{intra-basket retrieval} using the IC dataset, the model size of \textsc{METEOR} at ($K^a=10\%|A^a|,|C^a|=1\%|A^a|$) is similar to: \textsc{Dim Reduct} at ($d=25$); \textsc{Quantization} at ($4\;bit\;quant.$); \textsc{Hash Trick} at ($\gamma=10\%$); and \textsc{DCN+Hard Clustering} at ($\gamma=10\%$). Compared to the comparable models, \textsc{METEOR} outperforms the best baseline by $9.88\%$ in MRR and $17.55\%$ in $Recall@1$. This observation is consistent for both \textit{intra-basket item retrieval} and \textit{location retrieval}.

Out of the baselines, \textsc{Dim Reduct} and \textsc{Quantization} are the strongest baselines, but they have linearly increasing model sizes with respect to the total number of distinct units. Thus, the additional memory required to represent a newly seen unit is higher for those two baselines. For example, \textsc{Quantization} at ($4\;bit\;quant.$) requires 363\% more memory per unit compared to the corresponding \textsc{METEOR} at ($K^a=10\%|A^a|,|C^a|=1\%|A^a|$). The other two baselines: \textsc{Hash Trick}; and \textsc{DCN+Hard Clustering}, have significantly lower overheads for new units than \textsc{METEOR}. However, they perform a hard cluster assignment for each unit, which restricts the flexibility of representations, thus yielding poor performance for the downstream tasks.
In \textsc{DCN+Hard Clustering}, the hard clustering is performed incrementally at the end of each updating window. Subsequently, for a given unit, the embedding is taken as the centre of the cluster that the unit belongs to. However, as shown in~\cite{guo2017improved}, such hard assignment can degrade the embedding space, which could be the reason for having poor results with \textsc{DCN+Hard Clustering}. \textsc{Hash Trick} is the worst baseline, which could be due to the hard cluster assignment performed randomly without considering the semantics of the units. Hence, it is important to consider the semantics when learning memory efficient representations.

\textbf{Explicit Clusters as Noisy Fixed Clusters. }As we discussed in Section~\ref{subsubsec:step1}, \textsc{METEOR} can exploit the knowledge available in explicit grouping schemes when assigning the noisy fixed clusters for units. The results collected for \textit{intra-basket item retrieval} using such explicit product categories in CJ datasets are shown in Table~\ref{tab:explicit_clusters}.  As can be seen, the compressed embeddings in \textsc{METEOR} can even outperform \textsc{METERO-Full} while achieving $83\% (=(217MB-37MB)\times100\%/217MB)$ reduction in memory compared to \textsc{METERO-Full}, when the noisy fixed clusters are assigned using the product category information. This could be due to the shared basis vectors inside noisy fixed clusters. Thus, they can capture the additional knowledge introduced by explicit clusters too. This could help to improve the embeddings of rarely-occurring units (which have inaccurate embeddings in general) based on their neighbours in the same explicit cluster. To verify this point, the accuracy of the predicted labels for rarely occurring target test instances (appearing $< 10$ times) are examined. For those test instances, the compressed embeddings of \textsc{METEOR} with explicit noisy fixed clusters outperforms the costly embeddings of \textsc{METEOR} by nearly $2\%$ with respect to MRR. Thus, we can conclude that the compressed embeddings of \textsc{METEOR} with explicit noisy fixed clusters predicts rarely occurring ground truth instances more accurately than the costly embeddings.

\textbf{Ablation Study.} To check the significance of the role of the each element in \textsc{METEOR}, we perform an ablation study as shown in Table~\ref{tab:ablations_study}: (1) by removing \textit{weighted basis vector assignment}, which uniformly distributes basis vectors among noisy fixed clusters; and (2) by removing the \textit{sparsity constraint} in Equation~\ref{eq:compress}. As shown, the weighted basis vector assignment plays a significant role in \textsc{METEOR}. Although the sparsity constraint does not account for a significant performance improvement, we empirically observe that it yields compressed embeddings with many zeros, which could be exploited to reduce the memory to store the compressed embeddings using sparse matrix storage formats.    

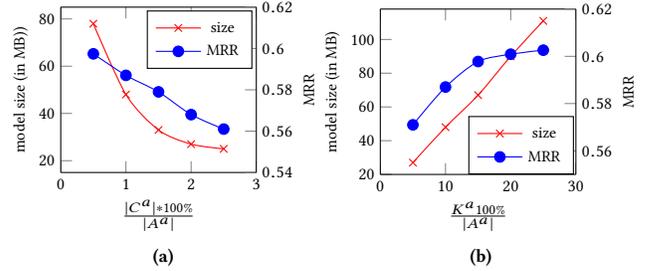
\begin{figure}[t]
\scriptsize
\centering
\subfloat[]{%
\begin{tikzpicture}
\label{fig:hyperpara_1}
\pgfplotsset{
    scale only axis,
    xmin=0, xmax=3
}

\begin{axis}[
  axis y line*=left,
  ymin=15, ymax=85,
  xlabel=
  $\frac{|C^a|*100\%}{|A^a|}$,
  ylabel=model size (in MB)),
  width = 2.6cm,
  height=2.2cm
]
\addplot[smooth,mark=x,red]
  coordinates{
    (0.5,78)
    (1,48)
    (1.5,33)
    (2, 27)
    (2.5, 25)
}; \label{plot_one}
\end{axis}

\begin{axis}[
  axis y line*=right,
  axis x line=none,
  ymin=0.54, ymax=0.62,
  ylabel=MRR,
  width = 2.6cm,
  height=2.2cm
]
\addlegendimage{/pgfplots/refstyle=plot_one}\addlegendentry{size}
\addplot[smooth,mark=*,blue]
  coordinates{
    (0.5,0.5974)
    (1,0.587)
    (1.5,0.579)
    (2, 0.568)
    (2.5, 0.561)
}; \addlegendentry{MRR}
\end{axis}
\end{tikzpicture}%
}
\subfloat[]{%
\begin{tikzpicture}
\label{fig:hyperpara_2}
\pgfplotsset{
    scale only axis,
    xmin=0, xmax=30
}

\begin{axis}[
  axis y line*=left,
  ymin=20, ymax=118,
  xlabel=$\frac{K^a100\%}{|A^a|}$,
  ylabel=model size (in MB),
  width = 2.6cm,
  height=2.2cm
]
\addplot[smooth,mark=x,red]
  coordinates{
    (5,27)
    (10,48)
    (15,67)
    (20,90)
    (25,111)
}; \label{plot_one}
\end{axis}

\begin{axis}[
  axis y line*=right,
  axis x line=none,
  ymin=0.55, ymax=0.62,
  ylabel=MRR,
width = 2.6cm,
  height=2.2cm,
  legend style={at={(0.45,0)},anchor=south west}
]
\addlegendimage{/pgfplots/refstyle=plot_one}\addlegendentry{size}
\addplot[smooth,mark=*,blue]
  coordinates{
    (5,0.571)
    (10,0.587)
    (15,0.5978)
    (20, 0.6009)
    (25, 0.6026)
}; \addlegendentry{MRR}
\end{axis}
\end{tikzpicture}
}
\vspace{-2mm}
\caption{Parameter Sensitivity on CJ: (a) MRRs for different $|C^a|$ values with $K^a=10\%|A^a|$; and (b) MRRs for different $K^a$ values with $|C^a|=1\%|A^a|$}\label{fig:results_hyperparameter}
\vspace{-6mm}
\end{figure}

\begin{figure*}[t]
\scriptsize
\centering
\subfloat[]{%
\begin{tikzpicture}
\pgfplotsset{
    scale only axis,
    xmin=0, xmax=6
}

\begin{axis}[
  axis y line*=left,
  ymin=2, ymax=11,
  xlabel=p,
  ylabel=Average time per record (in ms),
  width = 2.7cm,
  height=2.3cm
]
\addplot[smooth,mark=x,red]
  coordinates{
    (1,8.13)
    (2,7.44)
    (3, 6.28)
    (4, 5.41)
    (5, 4.87)
}; \label{plot_one}
\end{axis}

\begin{axis}[
  axis y line*=right,
  axis x line=none,
  ymin=0.560, ymax=0.610,
  ylabel=MRR,
  width = 2.7cm,
  height=2.3cm
]
\addlegendimage{/pgfplots/refstyle=plot_one}\addlegendentry{Avg time}
\addplot[smooth,mark=*,blue]
  coordinates{
    (1,0.587)
    (2,0.586)
    (3,0.585)
    (4,0.583)
    (5,0.580)
}; \addlegendentry{MRR}
\end{axis}
\end{tikzpicture}%
}
\subfloat[]{%
\begin{tikzpicture}
\pgfplotsset{
    scale only axis,
    xmin=0, xmax=6
}

\begin{axis}[
  axis y line*=left,
  ymin=2, ymax=11,
  xlabel=p,
  ylabel=Average time per record (in ms),
  width = 2.7cm,
  height=2.3cm
]
\addplot[smooth,mark=x,red]
  coordinates{
    (1,9.32)
    (2,7.98)
    (3,6.21)
    (4,4.87)
    (5,4.01)
}; \label{plot_one}
\end{axis}

\begin{axis}[
  axis y line*=right,
  axis x line=none,
  ymin=0.67, ymax=0.720,
  ylabel=MRR,
  width = 2.7cm,
  height=2.3cm
]
\addlegendimage{/pgfplots/refstyle=plot_one}\addlegendentry{Avg time}
\addplot[smooth,mark=*,blue]
  coordinates{
    (1,0.7084)
    (2,0.6999)
    (3,0.6975)
    (4,0.6941)
    (5,0.6922)
}; \addlegendentry{MRR}
\end{axis}
\end{tikzpicture}
}
\hspace{1em}
\subfloat[]{%
\begin{tikzpicture}
\pgfplotsset{
    scale only axis,
    xmin=0, xmax=6
}

\begin{axis}[
  axis y line*=left,
  ymin=8, ymax=17,
  xlabel=p,
  ylabel=Average time per record (in ms),
  width = 2.7cm,
  height=2.3cm
]
\addplot[smooth,mark=x,red]
  coordinates{
    (1, 15.2)
    (2, 13.9)
    (3, 12.2)
    (4, 10.7)
    (5, 9.8)
}; \label{plot_one}
\end{axis}

\begin{axis}[
  axis y line*=right,
  axis x line=none,
  ymin=0.74, ymax=0.79,
  ylabel=MRR,
  width = 2.7cm,
  height=2.3cm
]
\addlegendimage{/pgfplots/refstyle=plot_one}\addlegendentry{Avg time}
\addplot[smooth,mark=*,blue]
  coordinates{
    (1,0.7701)
    (2,0.7688)
    (3,0.7665)
    (4,0.7639)
    (5,0.7603)
}; \addlegendentry{MRR}
\end{axis}
\end{tikzpicture}%
}
\hspace{1em}
\subfloat[]{%
\begin{tikzpicture}
\pgfplotsset{
    scale only axis,
    xmin=0, xmax=6
}

\begin{axis}[
  axis y line*=left,
  ymin=8, ymax=18,
  xlabel=p,
  ylabel=Average time per record (in ms),
  width = 2.7cm,
  height=2.3cm
]
\addplot[smooth,mark=x,red]
  coordinates{
    (1,16.2)
    (2,14.9)
    (3, 12.0)
    (4, 10.9)
    (5, 9.9)
}; \label{plot_one}
\end{axis}

\begin{axis}[
  axis y line*=right,
  axis x line=none,
  ymin=0.75, ymax=0.80,
  ylabel=MRR,
  width = 2.7cm,
  height=2.3cm
]
\addlegendimage{/pgfplots/refstyle=plot_one}\addlegendentry{Avg time}
\addplot[smooth,mark=*,blue]
  coordinates{
    (1,0.7803)
    (2,0.7794)
    (3,0.7775)
    (4,0.7752)
    (5,0.7721)
}; \addlegendentry{MRR}
\end{axis}
\end{tikzpicture}
}
\vspace{-1mm}
\caption{Average time taken to process a record and MRR of METEOR-Parallel with $p$ number of parallel processors: (a) for \textit{intra-basket item retreival} using CJ; (b) for \textit{intra-basket item retreival} using IC; (c) for \textit{location retreival} using LA; and (d) for \textit{location retreival} using NY}\label{fig:results_meteor-parallel}
\end{figure*}
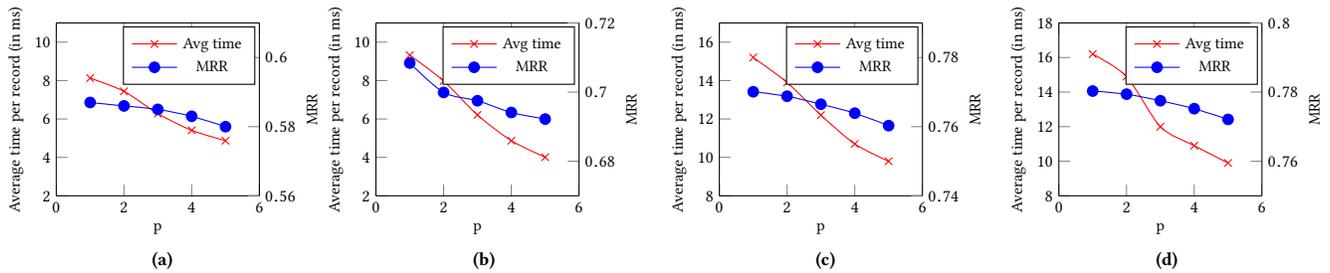

\textbf{Parameter Sensitivity.} In Figure~\ref{fig:results_hyperparameter}, we check the performance of \textsc{METEOR}, with different $K^a$ and $|C^a|$ values. For a given fixed $K^a$ value, when $|C^a|$ increases (see Figure\ref{fig:hyperpara_1}), the number of basis vectors per cluster reduces. Thus, the number of dimensions for compressed embedding reduces, in turn reducing the model size. Meanwhile, small noisy fixed clusters (with high $|C^a|$ values) increase the violations of Assumption 1 (see Section~\ref{subsubsec:step1}). This could be the reason for the performance drop when $|C^a|$ increases. As shown in Figure~\ref{fig:hyperpara_2}, when $K^a$ increases, both model sizes and MRRs for retrieval tasks monotonically increase and ultimately reach the model size and the performance of \textsc{METEOR-Full}. 

\textbf{(2) Online Learning in \textsc{METEOR}. }Table~\ref{tab:intra-basket} and Table~\ref{tab:location} also show the performance of the proposed online learning approach of \textsc{METEOR} (i.e., \textsc{METEOR-Full}) to update costly embeddings. Comparing \textsc{METEOR-Full} with the other online learning variants of \textsc{METEOR}: \textsc{METEOR-Info}; \textsc{METEOR-Decay}; and \textsc{METEOR-Cons}, \textsc{METEOR-Full}'s results are comparable (mostly superior) with sampling-based online learning variants of \textsc{METEOR}
(i.e., \textsc{METEOR-Decay} and \textsc{METEOR-Info}), which store historical records to avoid overfitting to recent records. Also, \textsc{METEOR-Full} outperforms \textsc{METEOR-Cons} as much as 30\% with respect to MRR, which has a similar memory complexity of \textsc{METEOR-Full}. Hence, the proposed adaptive optimization-based
online learning technique in \textsc{METEOR} achieves the performance of the state-of-the-art online learning methods (i.e., \textsc{METEOR-Decay} and \textsc{METEOR-Info}) in a memory-efficient manner without storing any historical records.

\textbf{(3) \textsc{METEOR-Parallel}. }Table~\ref{fig:results_meteor-parallel} shows the results collected with \textsc{METEOR-Parallel} with different numbers of parallel processes. For both retrieval tasks, \textsc{METEOR-Parallel} yields around $50\%$ reduction in the time taken to process a single record with 5 parallel processes. In contrast, the drop in the performance with respect to MRR with 5 processes are slight, which are 1.2\%, 2.3\%, 1.3\%, and 1.1\% for CJ, IC, LA, and NY respectively. Thus, \textsc{METEOR-Parallel} could be used to accelerate the embedding learning process of \textsc{METEOR}, while preserving the quality of the embeddings.

\section{Conclusion}
In this work, we proposed \textsc{METEOR}, which learns compact representations using multi-modal data streams in an online fashion. The learning of \textsc{METEOR} could be speeded up using parallel processes with different memory domains. Our results show that \textsc{METEOR} is a domain-agnostic framework, which can substantially reduce the memory complexity of conventional embedding learning approaches while preserving the quality of the embeddings. 

\textsc{METEOR} achieves around $80\%$ reduction in memory compared to the conventional costly embeddings without sacrificing performance by sharing parameters inside the semantically meaningful groupings of the multi-modal units. Hence, integrating \textsc{METEOR} with other similar explicit/implicit knowledge bases could be a promising research direction. Also, \textsc{METEOR} decides the number of shared parameters for each semantic group (i.e., noisy fixed clusters) based on their size. Besides, other factors could be considered (e.g., cluster shape) when assigning the basis vectors for noisy fixed clusters. In this work, we emulate the architecture of \textsc{METEOR-Parallel} using multi-threading, and we leave its actual implementation in a cluster of machines as future work. Also, we aim to explore the applications of \textsc{METEOR} for other domains.

\section*{Acknowledgement}
This research was financially supported by Melbourne Graduate Research Scholarship and Rowden White Scholarship.

\bibliographystyle{ACM-Reference-Format}
\bibliography{METEOR_Draft}

\end{document}